\documentclass{article}

\usepackage[preprint, nonatbib]{neurips_2022}
\usepackage{graphicx}
\usepackage{wrapfig}
\usepackage{xspace}

\newcommand{\name}{L2S\xspace}
\newcommand{\longname}{Language to Subtask\xspace}

\newcommand{\iglu}{IGLU\xspace}

\usepackage[utf8]{inputenc} 
\usepackage[T1]{fontenc}    
\usepackage{url}            
\usepackage{booktabs}       
\usepackage{amsfonts}       
\usepackage{nicefrac}       
\usepackage{microtype}      
\usepackage{xcolor}         

\title{Learning to Solve Voxel Building Embodied Tasks from Pixels and  Natural Language Instructions}

\author{%
Alexey Skrynnik \thanks{Corresponding author: \texttt{skrynnikalexey@gmail.com}} \thanks{AIRI} \footnotemark[3] \\  \and
Zoya Volovikova \thanks{MIPT}\\ \and
Marc-Alexandre Côté \thanks{Microsoft Research} \\ \and 
Anton Voronov \footnotemark[2] \\ \and
Artem Zholus \footnotemark[3]\\ \and
Negar Arabzadeh \thanks{University of Waterloo} \\ \and
Shrestha Mohanty \footnotemark[4] \\ \and
Milagro Teruel \footnotemark[4] \\ \and 
Ahmed Awadallah \footnotemark[4] \\ \and
Aleksandr Panov \footnotemark[2] \footnotemark[3] \\ \and
Mikhail Burtsev \footnotemark[2] \footnotemark[3] \\ \and
Julia Kiseleva \footnotemark[4] \\ \and
}

\begin{document}

    \maketitle

    \begin{abstract}
        The adoption of pre-trained language models to generate action plans for embodied agents is a promising research strategy. However, execution of instructions in real or simulated environments requires verification of the feasibility of actions as well as their relevance to the completion of a goal. We propose a new method that combines a language model and reinforcement learning for the task of building objects in a Minecraft-like environment according to the natural language instructions. Our method first generates a set of consistently achievable sub-goals from the instructions and then completes associated sub-tasks with a pre-trained RL policy.  The proposed method formed the RL baseline at the IGLU 2022 competition.
    \end{abstract}

    \section{Introduction}

    \begin{wrapfigure}{r}{0.6\textwidth}
        \centering
        \vspace{-1.11cm}
        \includegraphics[width=0.6\textwidth]{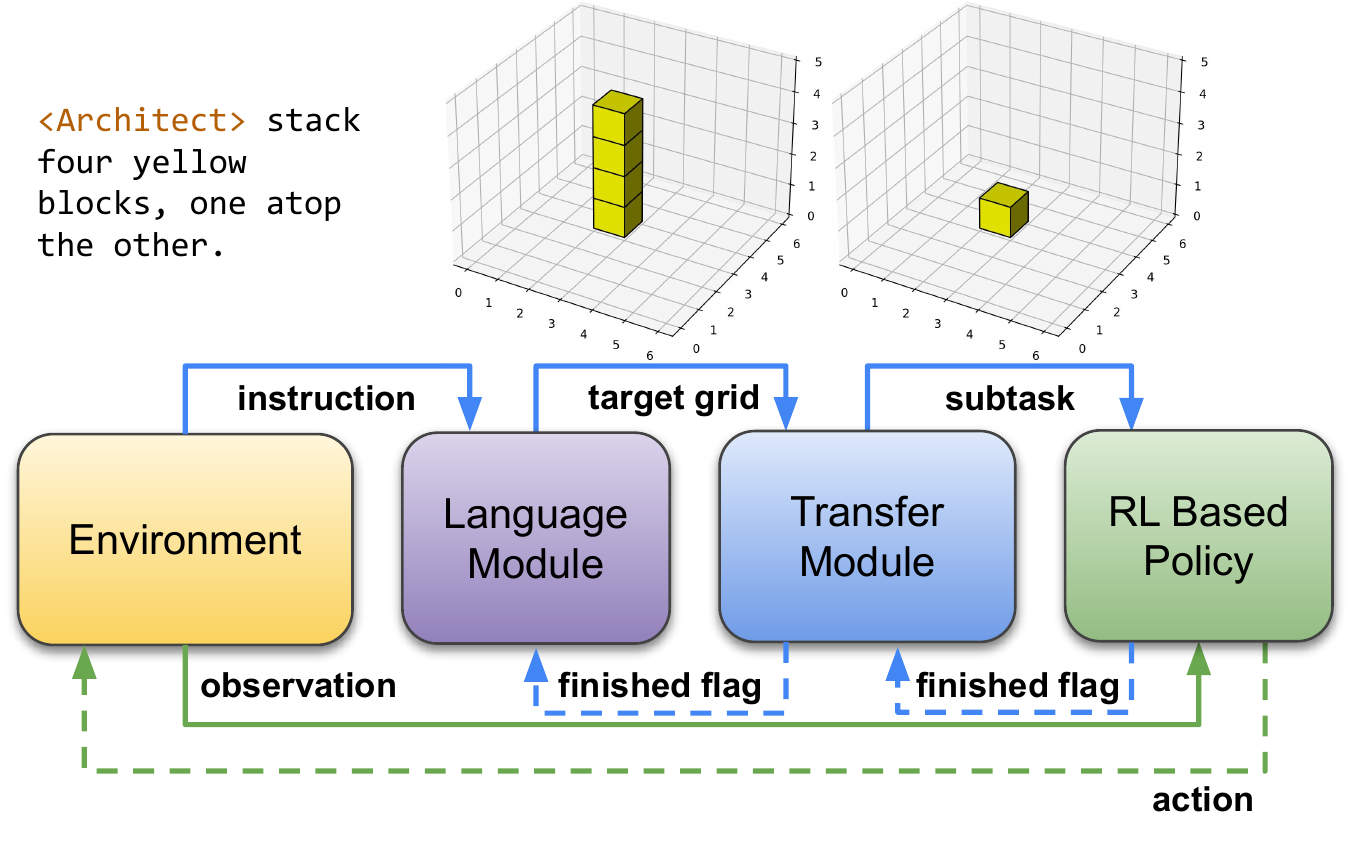}
        \caption{The general overview of the \longname Builder approach.}
        \vspace{-0.2 cm}
        \label{fig:scheme}
    \end{wrapfigure}

    The ability to learn complex actions in the environment with the help of knowledge expressed in natural language is a fundamental property of human intelligence. As well, the latest large language models (LLMs) can be considered as universal knowledge bases with which a human user can interact in natural language and solve quite complex tasks~\cite{huang2022language, zeng2022socratic}. It has been recently demonstrated that pre-trained LLMs are able to build an action plan in simulated and real environments~\cite{Bara2021,Min2021a,Murray2022}. Still, these LLM-based approaches require manual prompt engineering, handcrafted translation of language commands into embodied actions, and a strategy for goal-aware action selection from the distribution of possible options generated by language. In this regard, several studies~\cite{ahn2022can,ouyang2022training} show that the rough action plans extracted from the language model can be refined with reinforcement learning (RL). In this paper, we propose a method that leverages a pre-trained LM but requires no prompt engineering to learn a language-conditioned multitask RL policy for executing natural language instructions to build various figures in a Minecraft-like environment~\cite{pmlr-v176-kiseleva22a}.

    We use a simple embodied environment~\cite{Zholus2022b} proposed as a part of the \iglu competition~\footnote{https://www.iglu-contest.net/}, in which a \textit{Builder} agent should assemble a structure described by natural language instructions provided by an \textit{Architect} agent. We propose to solve this problem using Language to Subtask (\name) Builder consisting of three modules (fig. \ref{fig:scheme}). First, the \textit{Language} module learns to convert a natural language instruction into a sequence of commands to fill in the cells of the target structure's voxel representation. Then, the \textit{Transfer} module translates this sequence into a list of sub-goals to be executed one after another. Finally, in the \textit{Policy} module, a pre-trained RL-based policy is executed to complete each sub-goal while tracking the success of elementary actions.

    \section{Related Work}

    Action plans generation with language models becomes more and more actively explored ~\cite{Min2021a,huang2022language}. Some works focus on prompt engineering to properly control plan generation~\cite{zeng2022socratic}. While others pay attention to the problem of translating the output of a language model into feasible actions~\cite{huang2022language}. In \cite{ahn2022can}, models for selecting such feasible actions and executing elementary actions are trained with RL. In our work, we explore how to train a subtask completion policy conditioned on the next target state with RL.

    Building interactive embodied agents that learn to solve a task when provided with grounded natural language instructions in a collaborative environment is challenging. In the paper "Collaborative Dialogue in Minecraft"~\cite{narayan2019collaborative}, the authors provide a Minecraft-like environment and a dataset which consists of interactions between human Architect and Builder.
    The Architect knows the structure to be built and has to explain how to build it to the Builder using natural language. That work focused on developing and studying a model for predicting the architect's instructions.
    The task of predicting builder's action sequences, using only block placements and removals actions (no embodiment), is considered in the paper~\cite{jayannavar2020learning}. Human-guided collaborative problem-solving framework~\cite{kokel2022human} addresses the whole builder's problem. The paper proposes a nonparametric method that uses rule-based primitives and hierarchical task network (HTN) planners to build structures. In contrast to that approach, we propose a method to train a building policy, which utilizes modern natural language processing (NLP) and RL methods.

    \section{\longname Builder}
    \label{sec:baselines}
    
    Training an agent to build an arbitrary structure described in natural language is very challenging. Even if the task is given to the agent directly by specifying the target positions of elements on the grid, it is still nontrivial.  We propose an approach for training a general-purpose builder that can solve structures not seen during the training phase.

    The \longname Builder (\name) agent converts natural language description of a target structure to grid representation that defines subtasks sequentially completed by RL policy (see Figure~\ref{fig:scheme}). First, it predicts which blocks need to be placed from a given instruction. Then it executes a series of actions in the environment that lead to block placements according to the instruction. The builder agent consists of three modules: (1) the Language module which predicts coordinates and types of blocks given a text, (2) the rule-based Transfer module that iterates over positions of predicted blocks in a heuristically defined order, and (3) the RL Based Policy module that solves the atomic subtasks of one block placement or removal. The RL module operates with visual input, inventory, compass observation provided by the environment, and a target block position provided by the Transfer module.

    \subsection{\iglu Gridworld Environment}
    
    The \iglu Gridworld~\cite{Zholus2022b} environment represents the embodied agent with an ability to navigate, place, and break blocks of six different colors. The physics of the environment is very similar to the Minecraft game.
    The environment provides observation space including the agent's POV (RGB image), inventory state (available blocks of six different colors), and compass information. Additionally, we add into observation $3$D voxel ($3$D tensor), which encodes a subtask of adding or removing a block, predicted by the transfer module.

    In terms of action space, we opt for "walking" (as opposed to "flying"), which allows the agent to move with gravity enabled, jump, place, break, and select blocks from inventory.
    Moreover, we extend the agent's action space with one additional \textit{subtask completion} action. This new action enables feedback between the RL policy and the transfer module. If the agent selects this action the transfer module provides a new subtask at the next time step.

    Task evaluation uses $F_1$ score between the ``ground truth'', i.e. the target structure (a $3$D tensor), and the ``prediction'', i.e. the snapshot of the building zone at the end of the episode. The episode terminates either when the structure is complete, or when the time limit reaches $500$ time steps.
    
    \subsection{The Language Module}

    The Language module predicts block coordinates and block types autoregressively with fine-tuned T5-large (770 million parameters) encoder-decoder transformer~\cite{raffel2020exploring}. This model takes an instruction as an input and generates sparse block coordinates as an output.

    We consider the problem of generating the grid representation from instructions as an NLP problem of translating the Architect's instructions into a sequence of textual commands that the Builder can parse and interpret automatically. To ensure good natural language understanding we reuse T5 model a widely-used seq2seq Transformer, pre-trained to solve NLP tasks via text generation conditioned on the input prefix. For our instructions to commands task, we simply prepend all Architect's descriptions with the prefix: “implement given instructions: “ and fine-tuned the model to generate building commands in natural language.
    Initial data comes in the form of the dialog between the Architect and the Builder, so to distill instructions we have removed all builder’s utterances and concatenated the architect’s parts between different builder's actions into one sequence. For the output sequence, we have replaced blocks’ coordinates with their relative positions with respect to the starting block and concatenated consequent actions into one sequence. Next, the dataset was augmented with all possible permutations of colors of blocks.
    The biggest shortcoming of the described approach is that the model has no information about the current world state. In order to mitigate this issue, we add the last 3 input-output pairs to the current input during fine-tuning.  At the final stage, the sequence of commands generated by the model is converted into a voxel-based representation of the expected target structure.

    \subsection{The Transfer Module}

    \begin{wrapfigure}{r}{0.55\textwidth}
        \centering
        \includegraphics[width=0.55\textwidth]{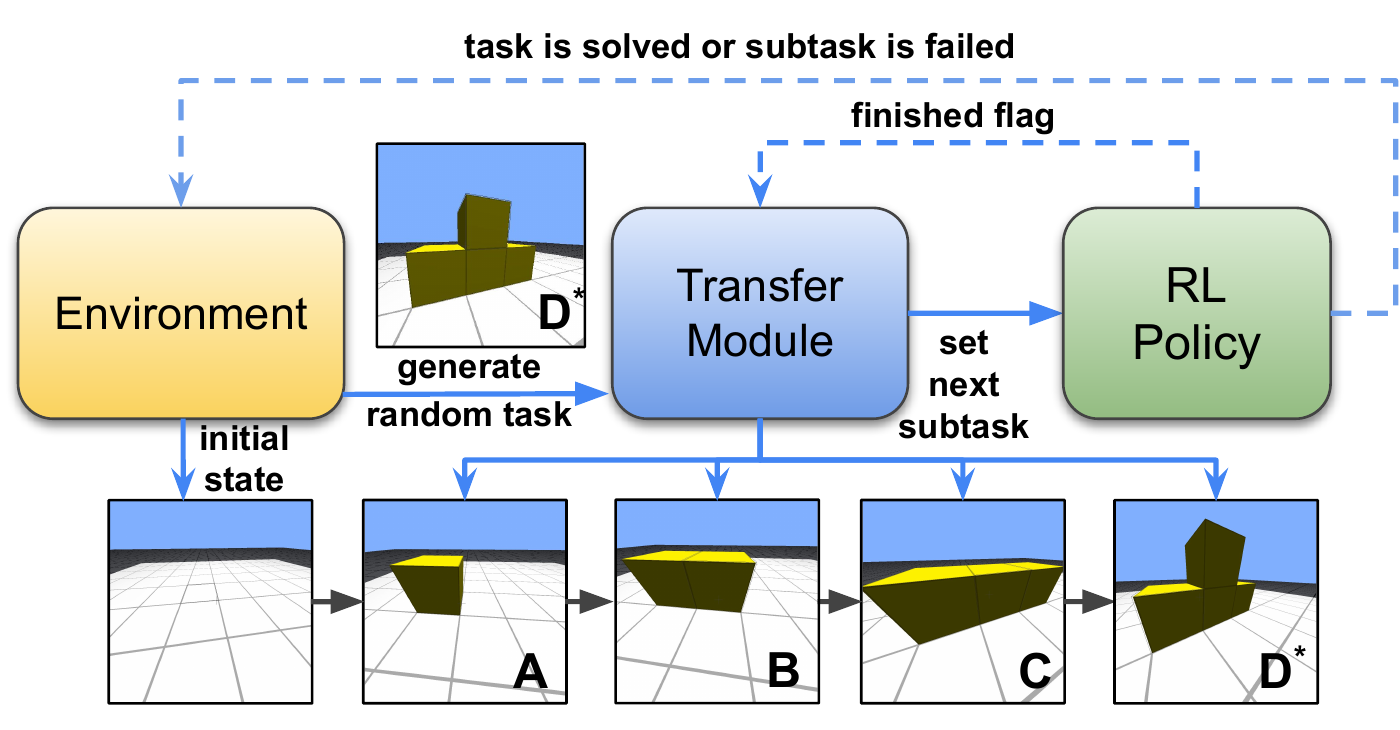}
        \caption{Scheme of the training process of the RL-based policy. The agent is trained to build any structure using a transfer module as a subtask generator and random structures as tasks. 
        }
        \label{fig:rl-training-scheme}
    \end{wrapfigure}  
    
    This module translates the voxel representation predicted by the Language module into a sequence of simple subtasks that require either placing or removing a single block. Subtasks are issued in a certain sequence, from bottom to top, from left to right, so when building a structure, the agent never encounters obstacles, which further simplifies the learning process. However, some structures may require building "flying" blocks (i.e., no supporting block below them). To handle that the agent is given a sequence of subtasks to build a full tower of blocks including the "flying one", and then an auxiliary sequence is given to build a tower for the agent to stand on and remove the unnecessary supporting blocks. Finally, the agent gets subtasks for removing the auxiliary tower.
    
    \subsection{The RL Based Policy Module}

    We use a high-throughput implementation of proximal policy optimization (PPO) for policy optimization. The RL agent is trained to build randomly generated compact structures. To encode the input the policy network uses a residual convolutional encoder, reproducing the backbone of IMPALA paper~\cite{espeholt2018impala} for processing RGB images. We use the same architecture removing max-pooling layers to encode the target block. The output of these encoders was concatenated with additional environment information (compass and inventory), and processed by multilayer perceptron (MLP) layers. The resultant vector is passed to the long short-term memory (LSTM) head.

    The reward function defines the reward signal based on the Manhattan distance of the placed block to the target position (see Table~\ref{table:reward-function}). In addition, the agent receives an extra reward, which is equal to $0.5$ if the agent puts the block in the right place, under his feet. This extra reward sufficiently increases the agent's performance in tasks with high structures.
    In order to teach the agent to provide subtask completion feedback, we added a reward equal to $1$ for correctly choosing the special \textit{subtask completion} action. In other cases, the agent is penalized by $-0.05$.

    \begin{table}[ht!]
        \centering
        \small
        \caption{The design of the reward function is based on Manhattan distance. The agent will receive a negative reward when the distance between the placed block and the target is more than three blocks. }
        \label{table:reward-function}
        \begin{tabular}{cccccccc}
            \toprule
            Distance & $0$ & $1$    & $2$    & $3$     & $4$       & $5$      & $i\in\{6, +\infty\}$ \\
            \midrule
            Reward   & $1$ & $0.25$ & $0.05$ & $0.001$ & $-0.0001$ & $-0.001$ & $-0.01 \times (i-5)$ \\
            \bottomrule
        \end{tabular}
        \vspace{-0.5cm}
    \end{table}
    
    \section{Experiments}

    In this section, we present the experimental results. For the training language module, we used a dataset from ~\cite{narayan2019collaborative}. The RL Policy is trained using only random structures.  We evaluate the whole pipeline using the open source dataset from \iglu competition~\cite{kiseleva2022iglu}.

    \begin{wrapfigure}{r}{0.4\textwidth}
        \centering
        \vspace{-1.33cm}
        \includegraphics[width=0.4\textwidth]{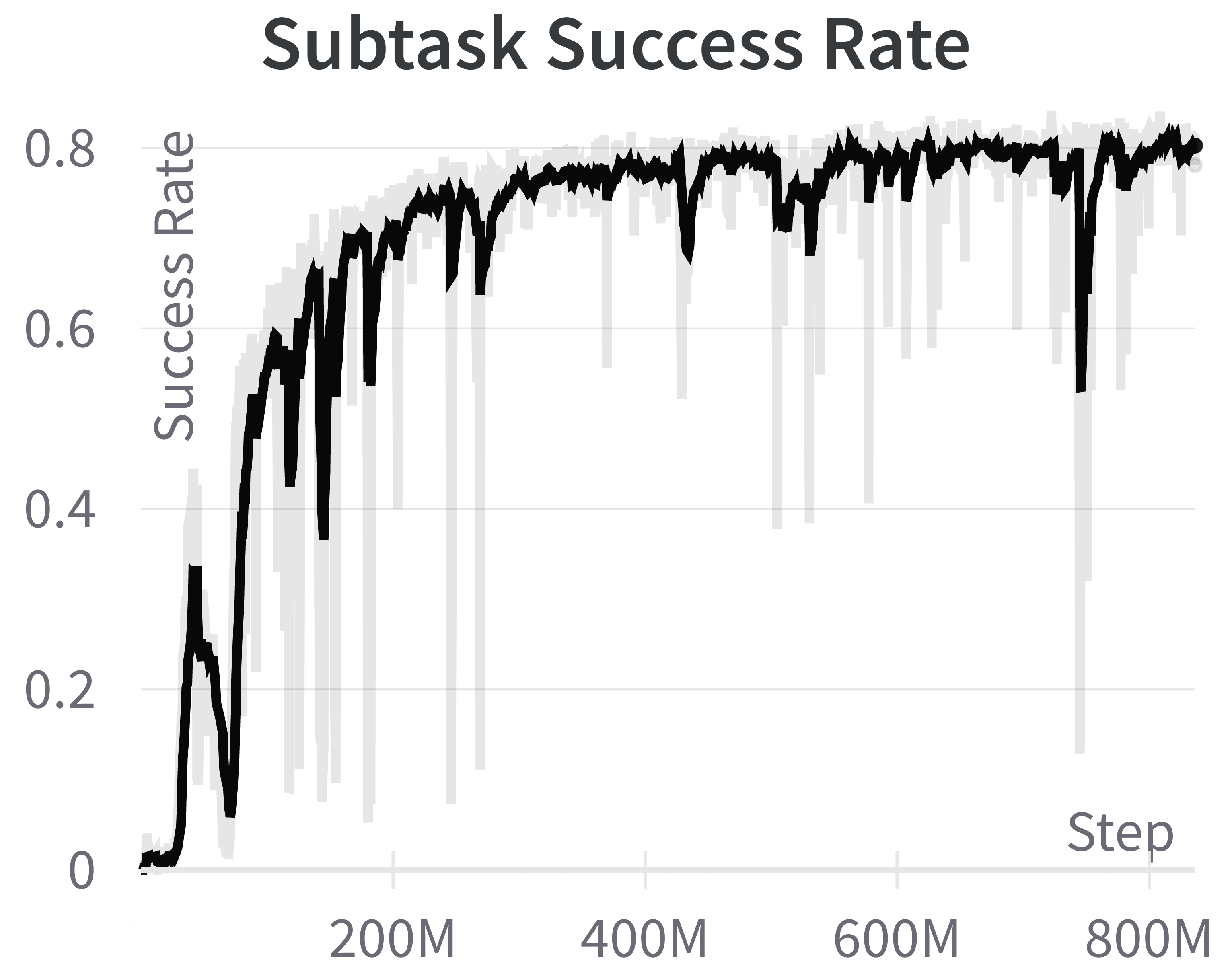}
        \caption{The average success rate of solving subtasks during training of RL policy.}
        \vspace{-0.2 cm}
        \label{fig:rl-training}
    \end{wrapfigure}
    
    \subsection{Training RL Agent}

    We trained RL policy for 1 billion environment steps using two Titan RTX GPUs (one for learning and one for sampling experience).
    Tasks for training the agent were generated by varying the parameters of the number of towers, their height, and the center of the building zone. We tried to cover as many possible sequences of subtasks as possible. The results for the training RL policy are presented in Figure~\ref{fig:rl-training}. We report the success rate metric, where an episode (subtask) is considered as successful if the agent placed the block in the correct place and then, chose the action to end the episode. The agent learns to solve the problem quite well, even given such a large space of possible subtasks.

    \subsection{Results}
    

    The results for the entire pipeline are presented in Table~\ref{table:final-results}. In addition to \name, we also report results of a perfect builder, which uses target tasks predicted by the language module. The results show that the perfect builder performs marginally better than the one trained using RL. The analysis of these episodes showed that the agent often cannot cope with complex structures with flying blocks, which were rarely generated during training. Also, the \name experiences problems when it is set in an environment with a partially built structure (tricky ones) that needs to be completed.
    
    \begin{table}[ht!]
        \centering
        \small
        \caption{Comparison of \name and a combination of the language module with the perfect builder. The tasks are labeled with the embodied skills required to solve them (based on \protect\cite{Zholus2022b} per skill aggregation).}
        \begin{tabular}{llllll|l}
            \toprule
        $F_1$ Score   & flat & flying & diagonal & tricky & tall &  all   \\
        \midrule
        \name Builder & 0.305 & 0.071 &  0.072 &  0.021 & 0.173 & 0.204   \\
        Perfect Builder & 0.428 & 0.118 & 0.089 & 0.104 & 0.366 & 0.270 \\
        \bottomrule
        \end{tabular}
        
        \label{table:final-results}
        \vspace{-0.5cm}
    \end{table}

    \section{Conclusion}
    
    In this paper, we proposed a new \longname Builder pipeline, which allows the agent to solve the problem of building arbitrary structures in the Minecraft-like environment according to language instructions. We propose an original transfer module that converts the coordinates of blocks predicted with fine-tuned language model into a sequence of subtasks for constructing the target object. To complete the subtasks of adding or removing blocks we pre-train a general purpose RL-based policy on a randomly generated dataset of subtasks. Our initial experiments demonstrate that the proposed \name Builder is able to consistently build multi-part objects described in natural language.  The resulting method can found its application as a baseline in the IGLU 2022 competition.

    \bibliographystyle{unsrt}
    \bibliography{main}

\begin{thebibliography}{10}

\bibitem{huang2022language}
Wenlong Huang, Pieter Abbeel, Deepak Pathak, and Igor Mordatch.
\newblock Language models as zero-shot planners: Extracting actionable
  knowledge for embodied agents.
\newblock {\em arXiv preprint arXiv:2201.07207}, 2022.

\bibitem{zeng2022socratic}
Andy Zeng, Adrian Wong, Stefan Welker, Krzysztof Choromanski, Federico Tombari,
  Aveek Purohit, Michael Ryoo, Vikas Sindhwani, Johnny Lee, Vincent Vanhoucke,
  et~al.
\newblock Socratic models: Composing zero-shot multimodal reasoning with
  language.
\newblock {\em arXiv preprint arXiv:2204.00598}, 2022.

\bibitem{Bara2021}
Cristian-Paul Bara, Sky CH-Wang, and Joyce Chai.
\newblock {MindCraft: Theory of Mind Modeling for Situated Dialogue in
  Collaborative Tasks}.
\newblock In {\em Proceedings of the 2021 Conference on Empirical Methods in
  Natural Language Processing}, pages 1112--1125, 2021.

\bibitem{Min2021a}
So~Yeon Min, Devendra~Singh Chaplot, Pradeep Ravikumar, Yonatan Bisk, and
  Ruslan Salakhutdinov.
\newblock {FILM: Following Instructions in Language with Modular Methods}.
\newblock In {\em ICLR}, 2022.

\bibitem{Murray2022}
Michael Murray and Maya Cakmak.
\newblock {Following Natural Language Instructions for Household Tasks With
  Landmark Guided Search and Reinforced Pose Adjustment}.
\newblock {\em IEEE Robotics and Automation Letters}, 7(3):6870--6877, jul
  2022.

\bibitem{ahn2022can}
Michael Ahn, Anthony Brohan, Noah Brown, Yevgen Chebotar, Omar Cortes, Byron
  David, Chelsea Finn, Keerthana Gopalakrishnan, Karol Hausman, Alex Herzog,
  et~al.
\newblock Do as i can, not as i say: Grounding language in robotic affordances.
\newblock {\em arXiv preprint arXiv:2204.01691}, 2022.

\bibitem{ouyang2022training}
Long Ouyang, Jeff Wu, Xu~Jiang, Diogo Almeida, Carroll~L Wainwright, Pamela
  Mishkin, Chong Zhang, Sandhini Agarwal, Katarina Slama, Alex Ray, et~al.
\newblock Training language models to follow instructions with human feedback.
\newblock {\em arXiv preprint arXiv:2203.02155}, 2022.

\bibitem{pmlr-v176-kiseleva22a}
Julia Kiseleva, Ziming Li, Mohammad Aliannejadi, Shrestha Mohanty, Maartje ter
  Hoeve, Mikhail Burtsev, Alexey Skrynnik, Artem Zholus, Aleksandr Panov, Kavya
  Srinet, Arthur Szlam, Yuxuan Sun, Katja Hofmann, Marc-Alexandre C{\^o}t{\'e},
  Ahmed Awadallah, Linar Abdrazakov, Igor Churin, Putra Manggala, Kata Naszadi,
  Michiel van~der Meer, and Taewoon Kim.
\newblock Interactive grounded language understanding in a collaborative
  environment: Iglu 2021.
\newblock In Douwe Kiela, Marco Ciccone, and Barbara Caputo, editors, {\em
  Proceedings of the NeurIPS 2021 Competitions and Demonstrations Track},
  volume 176 of {\em Proceedings of Machine Learning Research}, pages 146--161.
  PMLR, 06--14 Dec 2022.

\bibitem{Zholus2022b}
Artem Zholus, Alexey Skrynnik, Shrestha Mohanty, Zoya Volovikova, Julia
  Kiseleva, Artur Szlam, Marc-Alexandre Cot{\'{e}}, and Aleksandr~I. Panov.
\newblock {IGLU Gridworld: Simple and Fast Environment for Embodied Dialog
  Agents}.
\newblock In {\em CVPR 2022 Workshop on Embodied AI}, 2022.

\bibitem{narayan2019collaborative}
Anjali Narayan-Chen, Prashant Jayannavar, and Julia Hockenmaier.
\newblock Collaborative dialogue in minecraft.
\newblock In {\em Proceedings of the 57th Annual Meeting of the Association for
  Computational Linguistics}, pages 5405--5415, 2019.

\bibitem{jayannavar2020learning}
Prashant Jayannavar, Anjali Narayan-Chen, and Julia Hockenmaier.
\newblock Learning to execute instructions in a minecraft dialogue.
\newblock In {\em Proceedings of the 58th annual meeting of the association for
  computational linguistics}, pages 2589--2602, 2020.

\bibitem{kokel2022human}
Harsha Kokel, Mayukh Das, Rakibul Islam, Julia Bonn, Jon Cai, Soham Dan, Anjali
  Narayan-Chen, Prashant Jayannavar, Janardhan~Rao Doppa, Julia Hockenmaier,
  et~al.
\newblock Human-guided collaborative problem solving: A natural language based
  framework.
\newblock {\em arXiv preprint arXiv:2207.09566}, 2022.

\bibitem{raffel2020exploring}
Colin Raffel, Noam Shazeer, Adam Roberts, Katherine Lee, Sharan Narang, Michael
  Matena, Yanqi Zhou, Wei Li, Peter~J Liu, et~al.
\newblock Exploring the limits of transfer learning with a unified text-to-text
  transformer.
\newblock {\em J. Mach. Learn. Res.}, 21(140):1--67, 2020.

\bibitem{espeholt2018impala}
Lasse Espeholt, Hubert Soyer, Remi Munos, Karen Simonyan, Vlad Mnih, Tom Ward,
  Yotam Doron, Vlad Firoiu, Tim Harley, Iain Dunning, et~al.
\newblock Impala: Scalable distributed deep-rl with importance weighted
  actor-learner architectures.
\newblock In {\em International conference on machine learning}, pages
  1407--1416. PMLR, 2018.

\bibitem{kiseleva2022iglu}
Julia Kiseleva, Alexey Skrynnik, Artem Zholus, Shrestha Mohanty, Negar
  Arabzadeh, Marc-Alexandre C{\^o}t{\'e}, Mohammad Aliannejadi, Milagro Teruel,
  Ziming Li, Mikhail Burtsev, et~al.
\newblock Iglu 2022: Interactive grounded language understanding in a
  collaborative environment at neurips 2022.
\newblock {\em arXiv preprint arXiv:2205.13771}, 2022.

\end{thebibliography}





\end{document}